\documentclass[a4paper, 10pt, conference]{ieeeconf}      
\usepackage{FG2021}

\FGfinalcopy 

\IEEEoverridecommandlockouts                              
\title{\LARGE \bf
 Human Motion Prediction Using Manifold-Aware Wasserstein GAN
}

\author{\parbox{16cm}{\centering
    {\large Baptiste Chopin$^1$, Naima Otberdout$^1$, Mohamed Daoudi$^{2,3}$, Angela Bartolo$^4$}\\
    {\normalsize
    $^1$ Univ. Lille, CNRS, Centrale Lille, UMR 9189 CRIStAL, F-59000 Lille, France\\
    $^2$ IMT Lille Douai, Institut Mines-Télécom, Univ. Lille, Centre for Digital Systems, F-59000 Lille, France\\
    $^3$ Univ. Lille, CNRS, Centrale Lille, Institut Mines-Télécom, UMR 9189 CRIStAL, F-59000 Lille, France\\
    $^4$ Univ. Lille, CNRS, UMR 9193 SCALab, F-59000 Lille, France\\
    }}
    \thanks{This work was partially supported by the French State, managed by National Agency for Research (ANR) National Agency for Research (ANR) under the Investments for the future program with reference ANR-16-IDEX-0004 ULNE.}
}

\usepackage{times}
\usepackage{soul}
\usepackage{url}
\usepackage[hidelinks]{hyperref}
\usepackage[utf8]{inputenc}
\usepackage[small]{caption}
\usepackage{graphicx}
\usepackage{amsmath}

\usepackage{amsthm}

\usepackage{booktabs}
\usepackage[ruled,vlined]{algorithm2e}
\usepackage{algorithmic}

\usepackage{mathtools}
\usepackage{amssymb}
\usepackage{subfig}
\usepackage{color}

\newcommand{\std}[1]{\scriptscriptstyle\pm{#1}}
\DeclareMathOperator{\Tr}{tr}

\DeclarePairedDelimiter{\norm}{\lVert}{\rVert}

\begin{document}


\maketitle

\begin{abstract}

Human motion prediction aims to forecast future human poses given a prior pose sequence. The discontinuity of the predicted motion and the performance deterioration in long-term horizons are still the main challenges encountered in current literature. In this work, we tackle these issues by using a compact manifold-valued representation of human motion. Specifically, we model the temporal evolution of the 3D human poses as trajectory, what allows us to map human motions to single points on a sphere manifold. To learn these non-Euclidean representations, we build a manifold-aware Wasserstein generative adversarial model that captures the temporal and spatial dependencies of human motion through different losses. Extensive experiments show that our approach outperforms the state-of-the-art on CMU MoCap and Human 3.6M datasets. Our qualitative results show the smoothness of the predicted motions. The pretrained models and the code are provided at the following \href{https://drive.google.com/drive/folders/1pQkwtVDBeubW1oPwuXWFOOtftHabKaph}{\textcolor{magenta}{link}}.\\

\end{abstract}

\section{INTRODUCTION}
\begin{figure}[!ht]
    \centering
    \includegraphics[width=0.8\columnwidth,height=8.5cm]{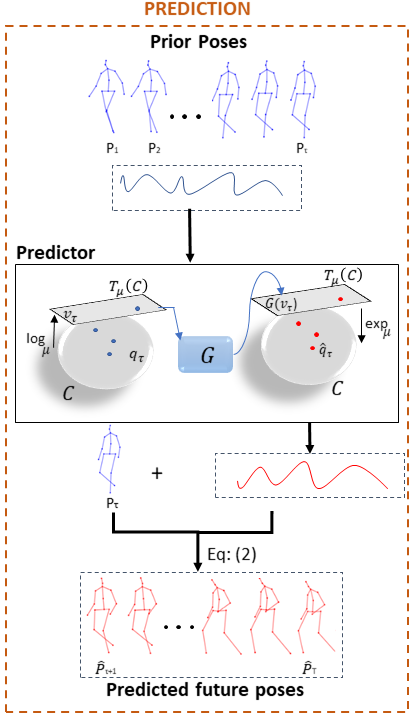}
    \caption{Overview of the human motion prediction process. Given a pose sequence history represented as a curve, then mapped to a single point in a hypersphere. The predictor maps the input point to a tangent space, then feeds it to the network $\mathcal{G}$ that predicts the future motion as a vector in $T_{\mu}(\mathcal{C})$. Exponential operator maps this vector to $\mathcal{C}$, before transforming it to a curve representing a motion. The predicted motion is transformed into a 3D human pose sequence corresponding to the future poses of the prior ones. }
    \label{fig:overview}
\end{figure}

 The problem of predicting future human motion is at the core of many applications in computer vision and robotics, such as human-robot interaction~\cite{KoppulaIEEEROS2013}, autonomous driving \cite{DBLP:journals/tiv/PadenCYYF16} and computer graphics~\cite{kovar2008motion}. In this paper, we are interested in building predictive models for short-term and long-term future 3D poses of a skeleton based on an initial history. Addressing this task gives rise to two major challenges: How to model the temporal evolution of the motion to ensure the smoothness of the predicted sequences? and how to take into consideration the spatial correlations between human joints to avoid implausible poses?  
\indent
Given the temporal aspect of the problem, human motion prediction was widely addressed with Recurrent Neural Networks (RNN)~\cite{DBLP:conf/iccv/FragkiadakiLFM15,jain_structural-rnn_2016,ghosh2017learning,martinez_human_2017}. However, while RNN based methods achieved good advance in term of accuracy, it was observed that the predicted motions present significant discontinuities due to the frame-by-frame regression process that  discourage the global smoothness of the motion. Besides, RNNs models accumulate errors across time, which results in large error and bad performance in long-term prediction. As a remedy, more recent works avoid these models and explore feed-forward networks instead. Including CNN~\cite{li_convolutional_2018}, GNN~\cite{mao2020learning} and fully-connected networks~\cite{butepage2017deep}, the hierarchical structure of feed-forward networks can better handle the spatial dependencies of human joints than RNNs. Nevertheless, these models require an additional strategy to encode the temporal information. To meet this challenge, an interesting idea was to model the human motion as trajectory~\cite{MaoICCV19}.\\
In this paper, we follow the idea of considering motions as trajectories but in a different context from the previous work. Among the advantages of our representation, the possibility to map these trajectories to single compact points on a manifold, which helps with the smoothness and the continuity of the predicted motions. In addition, the compact representation avoids the accumulation of errors through time and makes our method powerful for long-term prediction as illustrated in Figure~\ref{fig:introduction}. However, the resulting representations are manifold-valued data that cannot be handled with traditional generative models in a straightforward manner. 
To meet this challenge, we propose a manifold-aware Wasserstein Generative Adversarial Networks (WGAN) that anticipate future poses based on the input manifold-valued data that encodes the prior motion sequence. Our model incorporates the spatial dependencies between human joints through different loss functions that insure the plausibility of the predicted poses. A brief overview of our prediction process is illustrated in Figure~\ref{fig:overview}.\\
\textbf{Main contributions.} The paper gives rise to the following contributions: (1) To the best of our knowledge, this is the first approach that exploits compact manifold-valued representation for human motion prediction. By doing so, we model both the temporal and the spatial dependencies involved in human motion, resulting in smooth motions and plausible poses in long-term horizons. (2) We propose a manifold-aware WGAN for motion prediction. (3) Experimental results on Human 3.6M and the CMU MoCap datasets show quantitatively and visually the effectiveness of our method for short-term and long-term prediction.
.

\section{Related Work}\label{sec:RelatedWork}
\textbf{Human Motion Prediction with Deep Learning}. Since the problem of human motion prediction is a temporal dependent task, recurrent models were the first potential solution to be investigated, thus many works applied RNN and their variants to address this task. In~\cite{DBLP:conf/iccv/FragkiadakiLFM15}, the authors proposed a model that incorporates a nonlinear encoder and decoder before and after recurrent layers. Their approach was limited by the problem of error accumulation. Besides, they only capture the temporal dependencies while ignoring the spatial correlations between joints. To solve this issue, \cite{jain_structural-rnn_2016} proposed a Structural-RNN model relying on high-level spatio-temporal graphs. In an other direction, to reduce the effect of the error accumulation in recurrent models, \cite{ghosh2017learning} used a feed forward network for pose filtering and a RNN for temporal filtering. However, this strategy only reduces the accumulation of the error which still exists and affects the performance of recurrent models.
Taking a different direction, more recent works use feed-forward networks as an alternative model. To represent the temporal evolution with these models, different strategies were proposed. In ~\cite{li_convolutional_2018,butepage2017deep}, convolution across time was adopted to model the temporal dependencies with convolution networks, while~\cite{MaoICCV19} exploit Discrete Cosine Transform to encode the motion as trajectory. \\ In this paper, we take a completely different direction and we propose to deal with human motion by exploiting a manifold-valued representation with generative adversarial models. \\
\indent\textbf{Generative Adversarial Networks (GANs)}: Human motion prediction was also addressed with GANs in~\cite{ferrari_adversarial_2018} and~\cite{barsoum2018hp}, however, their generator is based on RNN structures to deal with the temporal aspect of this task. By doing so, their models keep the problem of error accumulation which may affect their performance in the long-term. In our work we completely discard recurrent models by adopting a compact representation of the human motion. \\ 
Motivated by the interest of manifold-valued images in a variety of applications, \cite{ZhiwuHuang2017} proposed manifold-aware WGAN. Inspired from this work, we build a manifold-aware WGAN that predict the future points of a poses trajectory given previous pose sequence. However, our model is different from the one proposed in~\cite{ZhiwuHuang2017} in two ways. Firstly, instead of unsupervised image generation from a vector noise, our model addresses the problem of predicting future manifold-valued representations from a manifold-valued inputs. Besides, we propose different objective functions to train our model on the task at hand.\\
\indent
\textbf{Modeling Human motions as trajectories on a Riemannian Manifold}: While our present work is the first that explores the benefit of manifold-valued trajectories  for human motion prediction, representing 3D human poses and their temporal evolution as trajectories on a manifold was adopted in many recent works for action recognition. 
Different manifolds were considered in different studies ~\cite{Turagacvpr2009}, ~\cite{Boulbaba2016PAMI}, ~\cite{KacemPAMI2020}. 
 More related to our work, in~\cite{devanne20143}, a human action is interpreted as a parametrized curve and is seen as a single point on the sphere by computing its Square Root Velocity Function (SRVF). Accordingly, different actions were classified based on the distance between their associated points on the sphere. All papers mentioned above show the effectiveness of  motion modeling as a trajectory in action recognition. Motivated by this fact, we show in this paper the interest of using such representation to address the recent challenges that still encountered in human motion prediction.
 
\section{Human Motion Modeling}\label{sec:motionModeling}
Two 3D skeleton representations were adopted for human motion prediction; angles based and 3D coordinates based representations. The first one models each joint by its rotation in term of Euler angles, while  the second representation uses the 3D coordinates of the joints. More recently,~\cite{mao2020learning}, showed in their experiments that the angles based representation where two different sets of angles can represent the exact same pose, leads to ambiguous results and cannot provide a fair and reliable comparison. Motivated by this, we use 3D joint coordinates to represent our skeleton poses.

\subsection{Representation of Pose Sequences as Trajectories in $\mathbb{R}^n$ }
Let $k$ be the number of joints that compose the skeleton, we represent $P_t$ the pose of the skeleton at frame $t$ by a n-dimensional tuple: $P_t=[x_1(t), y_1(t), z_1(t) \dots x_{k}(t), y_{k}(t), z_{k}(t)]^T ,$
\noindent the pose $P_t$ encodes the positions of $k$ distinct joints in $3$ dimensions. Consequently, an action sequence of length $T$ frames, can be described as a sequence $\{P_1, P_2 \dots ,P_T\}$, where $P_i  \in \mathbb{R}^n$ and $n=3 \times k$.

\noindent This sequence represents the evolution of the action over time and can be considered as a result of sampling a continuous curve in $\mathbb{R}^n$. Based on this consideration, we model in what follows, each pose sequence of a skeleton, as a continuous curve in $\mathbb{R}^n$ that describes the continuous evolution of the sequence over time. \\
Let us represent the curve describing a pose sequence by a continuous parameterized function $ \alpha (t): I =[0, 1] \rightarrow \mathbb{R}^n$. In this work, we formulate the problem of human motion prediction given the first consecutive frames of the action as the problem of predicting the possible next points of the curve describing these first frames. More formally, the problem of predicting the future poses $\left\{P_{\tau+1},P_{\tau+2}, \dots ,P_T \right\}$, given the first $\tau$ consecutive skeleton poses $\left\{P_1, P_2, \dots ,P_\tau \right\}$, where $\tau < T$, is formulated as the problem of predicting $ \alpha (t)_{t=\tau+1 \dots T}$ given $ \alpha (t)_{t=1 \dots \tau}$, such that, $ \alpha (t)$ is the continuous function representing the curve associated to the pose sequence $\left\{P_1,P_2, \dots ,P_T \right\}$.

\subsection{Representation of Human Motions as Elements in a Hypersphere $\mathcal{C}$ }
For the purpose of modeling and studying our curves, we adopt square-root velocity function (SRVF) proposed in~\cite{SrivastavaKJJ11}. It was successfully exploited for human action recognition~\cite{devanne20143}, 3D face recognition~\cite{drira20133d} and facial expression generation~\cite{OtberdoutPAMI2020}. Conveniently for us, this function maps each curve $\alpha(t)$ to one point in a hypersphere which provides a compact representation of the human motion. Specifically, for a given curve $ \alpha (t): I  \rightarrow \mathbb{R}^n$, the  square-root velocity function (SRVF) $ q(t): I\rightarrow \mathbb{R}^n$ is defined by the formula
\begin{equation}
q(t) = \frac{\dot{\alpha}(t)}{\sqrt{ \|\dot{\alpha}(t)\|}} \;,
\label{curve_2_q}
\end{equation}

\noindent where, $\|\cdot\|$ is the Euclidean 2-norm in $\mathbb{R}^n$. 
We can easily recover the curve (\emph{i.e}, pose sequence) $\alpha(t)$ from the generated SRVF (\emph{i.e}, dynamic information) $q(t)$ by,
\begin{equation}
\label{eq:curve_eq}
\alpha(t) = \int_0^t \norm{q(s)}q(s)ds + \alpha(0)\; ,
\end{equation}
\noindent where $\alpha(0)$ is the skeleton pose at the initial time step which corresponds in our case to the final time step of the history. 
In order to remove the scale variability of the curves, we scale them to be of length $1$. Consequently, the SRVF corresponding to these curves are elements of a unit hypersphere in the Hilbert manifold $\mathbb{L}^2(I,\mathbb{R}^{n})$ as explained in~\cite{SrivastavaKJJ11}. We will refer to this hypersphere as $\mathcal{C}$, such that, $
\mathcal{C}=\{q: \textit{I} \rightarrow \mathbb{R}^{n} | \; \|q\| = 1\} \subset \mathbb{L}^{2}(\textit{I},\mathbb{R}^{n})\;$. Each element of $\mathcal{C}$ represents a curve in $\mathbb{R}^n$ associated with a human motion. As $\mathcal{C}$ is a hypersphere, the geodesic length between two elements $q_1$ and $q_2$ is defined as: 

\begin{equation}
\label{eq:distance_sphere}
d_{\mathcal{C}}(q_1,q_2)=\cos^{-1}(\langle q_1,q_2 \rangle) \; . 
\end{equation}

\section{Architecture and Loss Functions }\label{sec:predictiveGAN}
Given a set of $m$ action sequences $\{ \{P_1,P_2, \dots P_T \}_i\}_{i=1}^m$ of $T$ consecutive skeleton poses. Let us consider the first $\tau$ poses $(\tau < T)$ as the actions history represented by their corresponding SRVFs $\{q_{\tau}^i\}_{i=1}^m$, and the last $(T-\tau)$ skeleton configurations as the future poses $\{q_{T}^i\}_{i=1}^m$ to be predicted.\\ Motivated by the success of generative adversarial networks, we aim to exploit these generative models to learn an approximation of the function $\Phi : \mathcal{C} \rightarrow \mathcal{C}$ that predicts the $(T-\tau)$ future poses from their associated $\tau$ prior ones. This can be achieved by learning the distribution of SRVFs data corresponding to future poses, on their underlying manifold \emph{i.e.}, hypersphere.  As stated earlier, SRVFs representations are manifold-valued data that cannot be used directly by classical GANs. This is due to the fact that the distribution of data having values on a manifold is quite different from the distribution of those lying on Euclidean space. 
\cite{ZhiwuHuang2017}, exploited the tangent space of the involved manifold and propose a manifold-aware WGAN that generates random data on a manifold. Inspired from this work, we propose a manifold-aware WGAN for motion prediction, to which we refer as PredictiveMA-WGAN, that can predict the future poses from the past ones. This is achieved by using the prior poses as input condition to the MA-WGAN. This condition is also represented by its SRVF; as a result PredictiveMA-WGAN takes manifold-valued data as input to predict its future, which is also a manifold-valued data. 

\subsection{Network Architecture}
PredictiveMA-WGAN consists of two networks trained in an adversarial manner: the predictor $\mathcal{G}$ and the discriminator $\mathcal{D}$. The first network $\mathcal{G}$ adjust its parameters to learn the distribution $\mathbb{P}_{q_T}$ of the future poses $q_T$ conditioned on the input prior ones $q_{\tau} $, while \textit{D} tries to distinguish between the real future poses $q_T$ and the predicted ones $\hat{q}_T$. 
During the training of these networks, we iteratively map the SRVF data back and forth to the tangent space using the exponential and the logarithm maps, defined in a particular point on the hypersphere. 

The predictor network is composed of multiple upsampling and downsampling blocks. It takes as input the prior poses $q_{\tau}$ and output the predicted future poses $\hat{q}_T$. A fully connected layer with $36864$ output channels and five upsampling blocks with $512$, $256$, $128$, $64$ and $1$ output channels, process the input prior pose. These upsampling blocks are composed of the nearest-neighbor upsampling followed by a $3 \times 3$ stride $1$ convolution and a Relu activation. The Discriminator \textit{D} contains three downsampling blocks with $64$, $32$ and $16$ output channels. Each block is a $3 \times 3$ stride $1$ Conv layer followed by batch normalization and Relu activation. These layers are then followed by two fully connected (FC) layers of $1024$ and $1$ outputs. The first FC layer uses Leaky ReLU and batch normalization. 

\subsection{Loss Functions}
In general, the objective of the training consists in minimizing the Wasserstein distance between the distribution of the predicted future poses $\mathbb{P}_{\hat{q}_{T}}$ and that of the real ones $\mathbb{P}_{q_T}$ provided by the dataset. Toward this goal we make use of the following loss functions:

\textbf{Adversarial loss} -- We propose an adversarial loss for predicting manifold-valued data from their history. The predictor takes a manifold-value data $q_{\tau}$ as input rather than a random vector as done in~\cite{ZhiwuHuang2017}, which requires to map these data to a tangent space using the logarithm map before feeding them to the network. Our adversarial loss is the following:
\begin{equation}
\begin{split}
\mathcal{L}_{a}= & {\mathbb{E}_{q_T \sim \mathbb{P}_{q_T}}\left[\mathcal{D}\left(\log _{\mu}(q_T)\right)\right]} 
\\ 
{ } & {-\mathbb{E}_{\mathcal{G}(log_\mu(q_{\tau})) \sim \mathbb{P}_{\hat{q}_T}}\left[\mathcal{D}\left(\log _{\mu}\left(\exp _{\mu}(\mathcal{G}(
\log _{\mu}
(q_{\tau})
))\right)\right)\right]}
\\ 
{ } & {+\lambda \mathbb{E}_{\widetilde{q} \sim \mathbb{P}_{\widetilde{q}}}\left[\left(\left\|\nabla_{\widetilde{q}} \mathcal{D}(\widetilde{q})\right\|-1\right)^{2}\right]},
\end{split}
\end{equation}

\noindent 
where $\log_\mu(.)$ and $\exp_\mu(.)$ are the logarithm and exponential maps on the sphere, used to iteratively map the SRVF data back and forth to the tangent space $T_{\mathcal{\mu}}(C)$ at a reference point $\mu$. They are given by:
\begin{equation}
\begin{split}
\label{Eq:LogSphereLog}
\log _{\mu}(q) &= \frac{d_{\mathcal{C}}(q,\mu)}{sin(d_{\mathcal{C}}(q,\mu))} (q - cos(d_{\mathcal{C}}(q,\mu))\mu) \, ,\\
\exp _{\mu}(s)&=cos(\|s\|)\mu + sin(\|s\|)\frac{s}{\|s\|},
 \end{split}
\end{equation}
\noindent 
where $d_{\mathcal{C}}(.,.)$ is the geodesic distance defined by~\eqref{eq:distance_sphere}.
The last term of $\mathcal{L}_a$ represents the gradient penalty proposed in~\cite{gulrajani2017improved}. $\widetilde{q}$ is a random sample following the distribution $\mathbb{P}_{\widetilde{q}}$, which is sampled uniformly along straight lines between pairs of points sampled from the real distribution $\mathbb{P}_{q_T}$ and the generated distribution $\mathbb{P}_{\hat{q}_T}$. It is given by:
$\label{Eq:eq}
\widetilde{q} = (1 - a) \log_\mu(q_T) + a\log_\mu(\exp_\mu(\mathcal{G}(\log _{\mu}(q_{\tau})))), $ where $\nabla_{\widetilde{q}}  D(\widetilde{q})$  is the gradient with respect to $\widetilde{q}$, and $0\leqslant a \leqslant 1 $.\\
The reference point $\mu$ of the tangent space used in our training is set to the mean of the training data. It is given by the Karcher mean~\cite{karcher1977riemannian} in $\mathcal{C}$, 
$\mu=\underset{{q_i} \in \mathcal{C}}{\operatorname{argmin}} \sum_{i=1}^m d_{\mathcal{C}}^2(\mu, q_i) $, where $\{q_i \}_{i=1}^m$ is $m$ training data.

\textbf{Reconstruction loss} -- In order to predict motions close to their ground truth, we add a reconstruction loss $\mathcal{L}_r$. This loss function quantifies the similarities in the tangent space $T_\mu(\mathcal{C})$ between the tangent vector $\log_\mu(q_{T})$ of the ground truth $q_T$ and its associated reconstructed vector $\log_\mu(\exp_\mu(\mathcal{G}(\log _{\mu}(q_{\tau}))))$. It is given by,
\begin{equation}
\label{eq:T_reconstruction_loss}
\mathcal{L}_r = \|{ \log_{\mu}(\exp_\mu({\mathcal{G}(\log _{\mu}(q_{\tau}))})) - \log_\mu(q_{T})} \|_1 \; ,
\end{equation}

\noindent
where $\|.\|_1$ denotes the $L_1$-norm. 

\textbf{Skeleton integrity loss} -- We propose a new loss function  $\mathcal{L}_{s}$ that minimizes the distance between the predicted poses and their ground truth as a remedy to the generation of abnormal skeleton poses.
 Indeed, the aforementioned loss functions rely only on the SRVF representations, which imposes constraints only on the dynamic information. However, to capture the spatial dependencies between joints that avoid implausible poses, we need to impose constraints on the predicted poses directly instead of their motions. By doing so, we predict dynamic changes that fit the initial pose and result in a long-term plausibility. The proposed loss function is based on the Gram matrix of the joint configuration $P$, $G = PP^T$, where $P$ can be seen as $k \times 3$ matrix. Let $G_i, G_j$  be two Gram matrices, obtained from joint poses $P_i, P_j \in \mathbb{R}^{k \times 3}$. The distance between $G_i$ and $G_j$ can be expressed \cite[p.~328]{GoluVanl96} as:
\begin{equation}
\label{eq:distW}
\Delta(G_i,G_j) = \Tr\left(G_i\right) + \Tr\left(G_j\right) - 2 \sum \limits_{{i=1}}^3 \sigma_{i} \; ,
\end{equation} 
\noindent where $\Tr(.)$ denotes the trace operator, and $\{\sigma_i\}_{i=1}^3$ are the singular values of $P_j^T P_i$.
\noindent
The resulting loss function is,

\begin{equation}
\label{eq:loss}
\mathcal{L}_{s} = \dfrac{1}{m}\dfrac{1}{\tau}\sum_{i=1}^{m}\sum_{t=1}^{\tau}\Delta (P_{i,t}, \hat{P}_{i,t})  \; ,
\end{equation}

\noindent where $m$ represents the number of training samples, $\tau$ is the length of the predicted sequence, $P$ is the ground truth pose and $\hat{P}$ is the predicted one.

\textbf{Bone length loss} -- To ensure the realness of the predicted poses, we impose further restrictions on the length of the bones. This is achieved through a loss function that forces the bone length to remain constant over time. Considering $b_{i,j,t}$  and $\hat{b}_{i,j,t}$ the $j$-th bones at time $t$ from the ground truth and the predicted $i$-th skeleton, respectively, we compute the following loss : 
\begin{equation}
\label{eq:loss_bone}
\mathcal{L}_{b} = \dfrac{1}{m}\dfrac{1}{\tau}\dfrac{1}{B}\sum_{i}^{m}\sum_{t=1}^{\tau}\sum_{j}^{B}\norm{b_{i,j,t}-\hat{b}_{i,j,t}}  \; ,
\end{equation}
with B the number of bones in the skeleton representation.

\textbf{Global loss} -- PredictiveMA-WGAN is trained using a weighted sum of the four loss functions $\mathcal{L}_a$, $\mathcal{L}_r$, $\mathcal{L}_s$ and $\mathcal{L}_b$ introduced above, such that,
\begin{equation}
\label{eq:PredictiveGAN_loss} 
\begin{array}{rl}
\mathcal{L}= {\beta_1 \mathcal{L}_{a} + \beta_2 \mathcal{L}_{r} + \beta_3 \mathcal{L}_{s} + \beta_4 \mathcal{L}_{b}}.
\end{array}
\end{equation}
\noindent
The parameters $\beta_i$ are the coefficients associated to different losses, they are set empirically in our experiments. 

The algorithm \ref{alg:WGANalgorithm} summarizes the main steps of our approach. It is divided in two stages, first we outline the steps needed to train our model, then we present the prediction stage, where the trained model is used to predict future poses of a given sequence. 
\begin{algorithm}[!ht]  
\tcp{Training}
  \KwData{$\{q_{\tau}^i\}_{i=1}^m$: SRVFs of training prior poses,
  $\{q_{T}^i\}_{i=1}^m$: real future poses,
  $\theta_0:$ initial parameters of $\mathcal{G}$, 
  $\eta_0:$ initial parameters of $\mathcal{D}$, 
  $\epsilon$: learning rate, $K$: batch size, $\lambda$: balance parameter of gradient penalty, $\zeta$: iterations number.
  }
 \KwResult{$\theta$: generator learned parameters.}
 \nl \For{$i=1 \dots \zeta$}{
 \nl Sample a mini-batch of $K$ random prior poses $\{q^j_{\tau} \}_{j=1}^K \sim \mathbb{P}_{q_{\tau}}$;
 
 \nl Sample a mini-batch of $K$ real future poses; $\{q^j_{T} \}_{j=1}^K \sim \mathbb{P}_{q_T}$;
 
 
 \nl $D_{\eta} \leftarrow{ \Delta_{\eta}(\mathcal{L})}, \mathcal{L}$ is given by Eq.~10;
 
 \nl $\eta \leftarrow \eta + \epsilon . AdamOptimizer(\eta, D_{\eta});$
 
\nl Sample a mini-batch of $K$ random prior poses; $\{q^j_{\tau} \}_{j=1}^K \sim \mathbb{P}_{q_\tau}$;
 
 \nl Compute $\{\mathcal{G}_{\theta}(\log _{\mu}(q_{\tau}^j))\}_{j=1}^K$;
 
\nl $G_{\theta} \leftarrow \Delta_{\theta}( -  D_{\eta}\left(\log _{\mu}\left(\exp _{\mu}(\mathcal{G}_{\theta}(\log _{\mu}(q_{\tau} ))\right)\right) ))$ 
 
\nl $\theta \leftarrow \theta + \epsilon . AdamOptimizer(\theta, G_{\theta});$}

\tcp{Prediction}
 \KwData{$\theta$: generator learned parameters, \\
 \hspace{0.8cm} $ \{ P_i\}_{i=1}^{\tau}$: Prior poses of a testing sequence. }
  \KwResult{$\{ \hat{P}_i\}_{i=\tau+1}^{T}$: Predicted future poses.}
  
 \nl Compute $q_{\tau}$ from $\{ P_i\}_{i=1}^{\tau}$ with Eq.~1;
 
 \nl Compute $\hat{q}_{T}= \exp_{\mu}( \mathcal{G}_{\theta}(\log _{\mu}(q_{\tau} )))$ using the learned parameters $\theta$;
 
 \nl Transform $\hat{q}_{T}$ into pose sequence $\{ \hat{P}_i\}_{i=\tau+1}^{T}$ using Eq.~2, with $\alpha(0) = P_{\tau} $

\caption{PredictiveMAWGAN algorithm}
\label{alg:WGANalgorithm}
\end{algorithm}
\section{Experiments}\label{sec:experiments}
In order to evaluate the proposed approach, we performed extensive experiments on two commonly used datasets. In what follows, we present and discuss our results.

\subsection{Datasets and Pre-processing}

\noindent \textbf{Human 3.6M}. Human 3.6M~\cite{ionescu_human36m_2014}  has 11 subjects in 15 various actions (Eating, Walking, Taking photos…). It is the largest dataset and the most commonly used for human motion prediction with 3D skeletons in literature. 
As previous works~\cite{martinez_human_2017,Cui_2020_CVPR}, our models are trained on $6$ subjects and tested on the specific clips of the $5$th subject. 
Following~\cite{Cui_2020_CVPR} we use only $17$ joints out of $32$; the removed joints correspond to duplicate joints, hands and feet. 

\noindent \textbf{CMU Motion Capture} (CMU MoCap). CMU Mocap dataset \footnote{\href{http://mocap.cs.cmu.edu}{http://mocap.cs.cmu.edu}} consists of $5$ categories, each containing several actions. To be coherent with~\cite{li_convolutional_2018}, we choose 8 actions: 'basketball', 'basketball signal', 'directing traffic','jumping', 'running', 'soccer', 'walking' and 'washing window'. We use the same joint configuration and pre-processing as for Human3.6M.

\subsection{Implementation Details}
 Our method is implemented using Tensorflow 2.2 on a PC with two 2.3Ghz processors, a Nvidia Quadro RTX 6000 GPU  and 64Go of RAM. The models are trained using the Adam optimizer \cite{KingmaICLR14}. The batch size is set to 64 and the number of epochs  is fixed to 500. The learning rate is fixed to $10^{-4}$. The loss coefficients $\beta_1$, $\beta_2$, $\beta_3$ and $\beta_4$ are respectively set to  1, 1, 10 and 10.

\subsection{Evaluation Metrics and Baselines}
We compare our results with state-of-the-art motion prediction methods that were based on 3D coordinate representation, including RNN based method (Residual sup).~\cite{martinez_human_2017}, CNN based method (ConvSeq2Seq)~\cite{li_convolutional_2018} and graph models; (FC-GCN) \cite{mao2020learning} and (LDRGCN)~\cite{Cui_2020_CVPR}.
We also compare with a simple baseline, Zero velocity introduced by~\cite{martinez_human_2017}, which sets all predictions to be the last observed pose at $t = \tau$. For LDRGCN we present the results reported by the authors for the method trained with data in 3D coordinate space.
For FC-GCN, ConvSeq2Seq and Residual sup., we present the results reported by \cite{mao2020learning} with the methods that use 3D coordinate data for training. For the long-term (1000ms) on Human 3.6M, we use the results presented by~\cite{Cui_2020_CVPR}  since they are not provided in \cite{mao2020learning}. We do not present the long-term results for Residual sup. on Human 3.6M as they are not available.

Following the state-of-the-art \cite{Cui_2020_CVPR}, our quantitative evaluation is based on the Mean Per Joint Position Error (MPJPE) \cite{ionescu_human36m_2014} in millimeter. This metric compares the predicted motions and their corresponding ground-truths in the 3D coordinate space. It is given by, 

 \begin{equation}
\label{eq:metric}
Err = \sqrt{\dfrac{1}{\Delta t}\dfrac{1}{k}\sum_{t=\tau+1}^{\tau+\Delta t}\sum_{j=1}^{k}\norm{p_{t,j}-\hat{p}_{t,j}}^2} \; ,
\end{equation}
where $p_{t,j} = [x_j(t),y_j(t),z_j(t)]$ are the coordinates of joint $j$ at time $t$ from the ground truth sequence, $\hat{p}_{t,j}$ the coordinates from the predicted sequence, $k$ the total number of joints in the skeleton, $\tau$ the number of frames in prior sequence and $\Delta t$ the number of predicted frames at which the sequence is evaluated.

\subsection{Quantitative Comparison}

In consistency with recent work, we report our results in short-term and long-term prediction. Given $10$ prior poses, $10$ future frames are predicted within $400$ms in short-term, while $25$ frames are predicted in $1$s for long-term prediction based on the previous $25$ frames. In Table~\ref{tab:average}, we compare our results with recent methods based on 3D joint coordinates representation. This latter, has been proven in~\cite{mao2020learning} to provide a reliable comparison in contrast to the angle based representation. The table shows a clear superiority of our approach over the state-of-the-art for both Human3.6M and CMU-MoCap datasets. We highlight that our approach in $80$ms and $160$ms is very competitive with LDRGCN approach, while in longer horizons we outperform this method in 320ms, 400ms and 1s, which demonstrates the robustness of our method in predicting motions that are closer to the ground-truth in long-term. \\

\begin{table}[!ht]
\centering
\small
\begin{tabular}{@{}l| c c c c c@{} } 
& \multicolumn{5}{c}{Human3.6M average}\\
millisecond (ms) & 80 & 160 & 320 & 400 & 1000 \\ \hline
Zero velocity& 19.6 & 32.5 & 55.1 & 64.4 & 107.9  \cr
Residual sup. & 30.8 & 57.0 & 99.8 & 115.5 & - \cr
convSeq2Seq & 19.6 & 37.8 & 68.1 & 80.3 & 140.5 \cr
FC-GCN & 12.2 & 25.0 & 50.0 & 61.3 & 114.7\cr
LDRGCN&  \textbf{10.7} & \textbf{22.5} & 43.1 & 55.8 & 97.8 \cr
Ours &  12.6 & \textbf{22.5} & \textbf{41.9} & \textbf{50.8} & \textbf{96.4} \cr \hline
& \multicolumn{5}{c}{CMU MoCap average}\\
millisecond (ms) & 80 & 160 & 320 & 400 & 1000 \\ \hline
Zero velocity& 18.4 &31.4  &56.2  & 67.7 &  130.5 \cr
Residual sup. &15.6 & 30.5 & 54.2 & 63.6 & 96.6 \cr
convSeq2Seq &12.5 & 22.2 & 40.7 & 49.7&84.6  \cr
FC-GCN & 11.5 & 20.4 & 37.8 & 46.8 & 96.5\cr 
LDRGCN& \textbf{9.4} & 17.6 & 31.6 & 43.1 & 82.9 \cr
Ours & \textbf{9.4} & \textbf{15.9} & \textbf{29.2} & \textbf{38.3} & \textbf{80.6} \cr \hline
\end{tabular}
\caption{\label{tab:average}Average error over all actions of Human3.6M and CMU MoCap. The short-term in $80$,$160$,$320$,$400ms$, and long-term in $1s$.}
\end{table}

We further report in Table~\ref{tab:HUMAN} and~\ref{tab:CMU}, our results and those of the literature on all actions of Human3.6M and CMU MoCap datasets, respectively. The protocol adopted by the baseline methods is to report the average error on eight randomly sampled test sequences. However, we found that the error is significantly affected by this random sampling, which makes it difficult to present a fair comparison. To alleviate this issue, we report the mean error obtained over $100$ runs; in each run, we randomly sample $8$ test sequences. Hence we report the average error as well as the standard deviation obtained with our model. Indeed, the standard deviation allows us to better measure the general performance of our model on different samples. According to Tables~\ref{tab:HUMAN} and~\ref{tab:CMU}, our approach outperforms the state-of-the-art especially for long-term prediction, which is consistent with the average error over all actions. Our results show also that the simple zero-velocity baseline outperforms the state of art in long-term for some actions (\emph{e.g,} Photo, Sitting and Walking dog for Human3.6H, Soccer and Jumping for CMU MoCap), while  in short-term, zero-velocity baseline error is generally higher. This evidences that the performance of the compared approaches decrease over time, while ours is more robust in long-term horizons, performing better than both the literature and the zero velocity baseline overall.

\begin{table*}[!ht]

\centering
\scriptsize\addtolength{\tabcolsep}{-3pt}
\begin{tabular}{l| p{0.725cm} p{0.725cm} p{0.725cm} p{0.725cm} p{0.725cm}| p{0.725cm} p{0.725cm} p{0.725cm} p{0.725cm} p{0.725cm}| p{0.725cm} p{0.725cm} p{0.725cm} p{0.725cm} p{0.725cm}| p{0.725cm} p{0.725cm} p{0.725cm} p{0.725cm} p{0.725cm} }

& \multicolumn{5}{c|}{Directions} & \multicolumn{5}{c|}{Discussion} & \multicolumn{5}{c|}{Eating} & \multicolumn{5}{c}{Greeting} \cr
millisecond (ms) & 80 & 160 & 320 & 400 & 1000 & 80 & 160 & 320 & 400 & 1000 & 80 & 160 & 320 & 400 & 1000 & 80 & 160 & 320 & 400 & 1000  \cr \hline
Zero velocity & 16.0 & 27.1 & 46.4 & 53.9 & 83.9 & 17.8 & 29.7 & 51.0 & 59.8 & 103.1 & 13.5 & 21.9 & 37.0 & 43.9 & 83.3 & 26.4 & 43.7 & 70.1 & 80.5 & \textbf{124.9}\cr
Residual sup.& 36.5 & 56.4 & 81.5 & 97.3 & - & 31.7 & 61.3 & 96.0 & 103.5 &- & 17.6 & 34.7 & 71.9 & 87.7 & - & 37.9 & 74.1 & 139.0 & 158.8& - \cr
convSeq2Seq& 22.0 & 37.2 & 59.6 & 73.4 & 118.3 & 18.9 & 39.3 & 67.7 & 75.7 & 123.9 & 13.7 & 25.9 & 52.5 & 63.3 & 74.4 & 24.5 & 46.2 & 90.0 & 103.1 & 191.2 \cr
FC-GCN  & 12.6 & 24.4 & 48.2 & 58.4 & 89.1 & 9.8 & 22.1 & 39.6 & 44.1 & 78.5 & 8.8 & 18.9 & 39.4 & 47.2 & 57.1 & 14.5 & 30.5 & 74.2 & 89.0 & 148.4  \cr
LDRGCN & 13.1 & 23.7 & 44.5 & 50.9 & \textbf{78.3} & \textbf{9.4} & \textbf{20.3} & \textbf{35.2} & \textbf{41.2} & \textbf{67.4} & \textbf{7.6} & \textbf{15.9} & 37.2 & 41.7 & \textbf{53.8} & \textbf{9.6} & \textbf{27.9} & 66.3 & 78.8 & 129.7  \cr
Ours & \textbf{11.1} & \textbf{20.9} & \textbf{38.8} & \textbf{47.0} & \textbf{83.5} & 11.9 & \textbf{22.7} & 44.8 & 54.6 & 102.2& \textbf{9.0} & \textbf{15.9} & \textbf{29.1} & \textbf{35.0} & 65.3 & 19.6 & 35.1 & \textbf{64.0} & \textbf{78.2} & \textbf{126.8}\cr
& $\boldsymbol{\std{2.7}}$ & $\boldsymbol{\std{4.9}}$ & $\boldsymbol{\std{8.4}}$ & $\boldsymbol{\std{9.7}}$ & $\boldsymbol{\std{15.3}}$ & $\std{1.9}$ & $\boldsymbol{\std{3.4}}$ & $\std{6.5}$ &$\std{7.7}$ & $\std{16.5}$ & $\boldsymbol{\std{1.5}}$ & $\boldsymbol{\std{2.8}}$ & $\boldsymbol{\std{4.8}}$ & $\boldsymbol{\std{5.3}}$ & $\std{6.8}$ & $\std{3.4}$ & $\std{6.8}$ & $\boldsymbol{\std{13.1}}$ & $\boldsymbol{\std{16.1}}$ & $\boldsymbol{\std{16.7}}$ \cr  \hline \hline

&  \multicolumn{5}{c|}{Phoning} &  \multicolumn{5}{c|}{Photo} & \multicolumn{5}{c|}{Posing} & \multicolumn{5}{c}{Purchase} \cr  
millisecond (ms) & 80 & 160 & 320 & 400 & 1000 & 80 & 160 & 320 & 400 & 1000 & 80 & 160 & 320 & 400 & 1000 & 80 & 160 & 320 & 400 & 1000 \cr \hline
Zero velocity &  15.8 & 26.5 & 43.7 & 51.0 & 92.3 & 16.9 & 28.4 & 49.2 & 58.3 & \textbf{98.8} & 20.4 & 34.7 & 61.5 & 73.3 & 136.1 & 22.1 & 36.5 & 61.8 & 72.2 & 126.3\cr
Residual sup. & 25.6 & 44.4 & 74.0 & 84.2 & - & 23.6 & 47.4 & 94.0 & 112.7 & - & 27.9 & 54.7 & 131.3 & 160.8 & - & 40.8 & 71.8 & 104.2 & 109.8 & - \cr
convSeq2Seq& 17.2 & 29.7 & 53.4 & 61.3 & 127.5 & 14.0 & 27.2 & 53.8 & 66.2 & 151.2 & 16.1 & 35.6 & 86.2 & 105.6 & 163.9 & 29.4 & 54.9 & 82.2 & 93.0 & 139.3  \cr
FC-GCN & 11.5 & 20.2 & 37.9 & 43.2 & 94.3 & \textbf{6.8} & 15.2 & 38.2 & 49.6 & 125.7 & 9.4 & 23.9 & 66.2 & 82.9 & 143.5 & 19.6 & 38.5 & 64.4 & 72.2 & 127.2 \cr
LDRGCN& \textbf{10.4} & \textbf{14.3} & \textbf{33.1} & \textbf{39.7} & 85.8 & 7.1 & \textbf{13.8} & \textbf{29.6} & 44.2 & 116.4 & \textbf{8.7} & \textbf{21.1} & 58.3 & 81.9 & \textbf{133.7} & 16.2  & 36.1 & 62.8 & 76.2 & \textbf{112.6}  \cr 
Ours & \textbf{11.7} & 19.4& \textbf{34.9} & \textbf{42.3} & \textbf{81.8} & \textbf{8.8} & \textbf{16.0} & \textbf{32.4} & \textbf{40.9} & \textbf{98.9} & 13.7 & \textbf{25.9} & \textbf{50.0} & \textbf{61.1} & \textbf{137.7} & \textbf{14.2} & \textbf{26.5} & \textbf{48.3} & \textbf{58.1} & \textbf{120.8}  \cr
& $\boldsymbol{\std{2.2}}$ & $\std{3.6}$ & $\boldsymbol{\std{6.4}}$ & $\boldsymbol{\std{7.6}}$ & $\boldsymbol{\std{9.8}}$ & $\std{2.0}$ & $\boldsymbol{\std{3.5}}$ & $\boldsymbol{\std{6.9}}$ & $\boldsymbol{\std{8.6}}$ & $\boldsymbol{\std{16.1}}$ & $\std{3.3}$ & $\boldsymbol{\std{6.3}}$ & $\boldsymbol{\std{11.0}}$ & $\boldsymbol{\std{12.7}}$ & $\boldsymbol{\std{12.8}}$ & $\boldsymbol{\std{2.5}}$ & $\boldsymbol{\std{4.8}}$ & $\boldsymbol{\std{9.8}}$ & $\boldsymbol{\std{12.5}}$ & $\boldsymbol{\std{19.0}}$ \cr  \hline \hline

& \multicolumn{5}{c|}{Sitting} & \multicolumn{5}{c|}{Sitting Down} & \multicolumn{5}{c|}{Smoking} & \multicolumn{5}{c}{Waiting} \cr   
millisecond (ms) & 80 & 160 & 320 & 400 & 1000 & 80 & 160 & 320 & 400 & 1000 & 80 & 160 & 320 & 400 & 1000 & 80 & 160 & 320 & 400 & 1000 \cr \hline
Zero velocity & 14.6 & 23.9 & 40.9 & 48.4 & \textbf{94.7} & 19.5 & 32.4 & 53.5 & 61.8 & \textbf{112.2} & 14.9 & 24.6 & 41.7 & 49.3 & 84.0 & 17.0 & 28.2 & 48.9 & 57.8 & 99.4\cr
Residual sup.& 34.5 & 69.9& 126.3 & 141.6 & - & 28.6 & 55.3 & 101.6 & 118.9 & - & 19.7 & 36.6 & 61.8 & 73.9 & - & 29.5 & 60.5& 119.9 & 140.6& -\cr
convSeq2Seq& 19.8 & 42.4 & 77.0 & 88.4 & 132.5 & 17.1 & 34.9 & 66.3 & 77.7 & 177.5 & 11.1 & 21.0 & 33.4 & 38.3 & 52.2 & 17.9 & 36.5 & 74.9 & 90.7 & 205.8 \cr
FC-GCN & 10.7 & 24.6 & 50.6 & 62.0 & 119.8 & 11.4 & 27.6 & 56.4 & 67.6 & 163.9 & 7.8 & 14.9 & 25.3 & 28.7 & 44.3 & 9.5 & 22.0 & 57.5 & 73.9 & 157.2 \cr
LDRGCN& \textbf{9.2} & 23.1 & 47.2 & 57.7 & 106.5 & \textbf{9.3} & \textbf{21.4} & \textbf{46.3} & \textbf{59.3} & 144.6 & 8.1 & \textbf{13.4} & \textbf{24.8} & \textbf{24.9} & \textbf{43.1} &  \textbf{9.2} & \textbf{17.6} & 47.2 & 71.6 & 127.3 \cr
Ours & \textbf{10.4}& \textbf{17.9} & \textbf{33.1} & \textbf{40.7} & \textbf{97.7}& 15.8& 28.2& \textbf{52.9} & \textbf{64.5}& \textbf{125.2}& \textbf{7.9} & \textbf{14.3}& \textbf{25.2}& \textbf{30.4}& 63.4& \textbf{11.4} & \textbf{20.3} & \textbf{38.8} & \textbf{47.2} & \textbf{94.0}\cr
& $\boldsymbol{\std{2.8}}$& $\boldsymbol{\std{3.5}}$& $\boldsymbol{\std{5.3}}$& $\boldsymbol{\std{6.4}}$ & $\boldsymbol{\std{14.0}}$ & $\std{3.4}$& $\std{5.1}$ & $\boldsymbol{\std{9.3}}$ & $\boldsymbol{\std{11.5}}$ & $\boldsymbol{\std{23.3}}$ & $\boldsymbol{\std{1.6}}$& $\boldsymbol{\std{2.7}}$ & $\boldsymbol{\std{4.5}}$ & $\boldsymbol{\std{5.2}}$ & $\std{9.7}$ & $\boldsymbol{\std{3.1}}$ & $\boldsymbol{\std{4.3}}$ & $\boldsymbol{\std{7.6}}$ & $\boldsymbol{\std{9.0}}$ & $\boldsymbol{\std{13.7}}$ \cr  \hline \hline

& \multicolumn{5}{c|}{Walking Dog} & \multicolumn{5}{c|}{Walking} & \multicolumn{5}{c|}{Walking Together} & \multicolumn{5}{c}{Average}\cr 
millisecond (ms) & 80 & 160 & 320 & 400 & 1000 & 80 & 160 & 320 & 400 & 1000 & 80 & 160 & 320 & 400 & 1000 &80 &160 & 320 & 400 & 1000 \cr \hline 
Zero velocity & 26.9 & 42.3 & 69.2 & 79.5 & 119.2 & 28.1 & 49.2 & 86.0 & 100.3 & 149.1 & 23.5 & 39.2 & 65.4 & 75.6 & 111.3 & 19.6 & 32.5 & 55.1 & 64.4 & 107.9  \cr
Residual sup.& 60.5 & 101.9 & 160.8 & 188.3 & - & 23.8 & 40.4 & 62.9 & 70.9 & - & 23.5 & 45.0 & 71.3 & 82.8 & - & 30.8 & 57.0 & 99.8 & 115.5 & - \cr
convSeq2Seq& 40.6 & 74.7 & 116.6 & 138.7 & 210.2 & 17.1 & 31.2 & 53.8 & 61.5 & 89.2 & 15.0 & 29.9 & 54.3 & 65.8 & 149.8 & 19.6 & 37.8 & 68.1 & 80.3 & 140.5 \cr
FC-GCN & 32.2 & 58.0 & 102.2 & 122.7 & 185.4 & \textbf{8.9} & 15.7 & 29.2 & 33.4 & 50.9 & 8.9 & 18.4 & 35.3 & 44.3 & 102.4 & 12.2 & 25.0 & 50.0 & 61.3 & 114.7\cr
LDRGCN& 25.3 & 56.6 & 87.9 & 99.4 & 143.2 & \textbf{8.9} & \textbf{14.9} & \textbf{25.4} & \textbf{29.9} & \textbf{45.8} & \textbf{8.2} & \textbf{18.1} & \textbf{31.2} & \textbf{39.4} & 79.2 & \textbf{10.7} & \textbf{22.5} & 43.1 & 55.8 & 97.8 \cr
Ours & \textbf{19.3} & \textbf{34.2} & \textbf{65.6} & \textbf{77.5} & \textbf{117.8} & 12.0 & 21.1& 35.6 & 42.4 & 68.2 & 11.6 & \textbf{19.7} & 34.5 & \textbf{41.8} &\textbf{63.4} & 12.6 & \textbf{22.5} & \textbf{41.9} & \textbf{50.8} & \textbf{96.4} \cr
& $\boldsymbol{\std{5.9}}$ & $\boldsymbol{\std{9.5}}$ & $\boldsymbol{\std{17.8}}$ & $\boldsymbol{\std{19.7}}$ & $\boldsymbol{\std{23.7}}$ & $\std{1.1}$ & $\std{1.7}$ & $\std{2.9}$ & $\std{3.8}$ & $\std{5.3}$ & $\std{1.1}$ & $\boldsymbol{\std{1.6}}$ & $\std{3.0}$ & $\boldsymbol{\std{3.8}}$ & $\boldsymbol{\std{6.4}}$  \cr  \hline

\end{tabular}
\caption{\label{tab:HUMAN} Motion prediction results measured with eq.\ref{eq:metric} for all actions in the Human 3.6M dataset for short-term within $80$, $160$, $320$, $400$ms, and long-term in $1$s. Best results in bold, while state-of-the-art best results that fit in our confidence interval are also written bold.}
\end{table*}

\begin{table*}[!ht]

\centering
\scriptsize\addtolength{\tabcolsep}{-3pt}
\begin{tabular}{@{}l| p{0.725cm} p{0.725cm} p{0.725cm} p{0.725cm} p{0.725cm}| p{0.725cm} p{0.725cm} p{0.725cm} p{0.725cm} p{0.725cm}| p{0.725cm} p{0.725cm} p{0.725cm} p{0.725cm} p{0.725cm}| p{0.725cm} p{0.725cm} p{0.725cm} p{0.725cm} p{0.725cm} @{} }

& \multicolumn{5}{c|}{Basketball} & \multicolumn{5}{c|}{Basketball signal} & \multicolumn{5}{c|}{Directing traffic} & \multicolumn{5}{c}{Jumping} \cr
millisecond (ms) & 80 & 160 & 320 & 400 & 1000 & 80 & 160 & 320 & 400 & 1000 & 80 & 160 & 320 & 400 & 1000 & 80 & 160 & 320 & 400 & 1000 \cr \hline
Zero velocity & 20.3 & 34.6 & 62.2& 75.0 & 143.5 & 6.4 & 11.0 & 19.9 & 24.2 & 50.5 & 26.6 & 41.9 & 69.1 & 81.9 &155.3 & 21.4 & 36.3 & 63.2 & 75.2 &138.8 \cr
Residual sup. & 18.4 & 33.8 & 59.5 & 70.5 & 106.7 & 12.7 & 23.8 & 40.3 & 46.7 & 77.5 & 15.2 & 29.6 & 55.1 & 66.1 & 127.1 & 36.0 & 68.7 & 125.0 & 145.5 & 195.5 \cr
convSeq2Seq & 16.7 & 30.5 & 53.8 & 64.3 & \textbf{91.5} & 8.4 & 16.2 & 30.8 & 37.8 & 76.5 & 10.6 & 20.3 & 38.7 & 48.4 & \textbf{115.5} & 22.4 & 44.0 & 87.5 & 106.3 & 162.6 \cr
FC-GCN & 14.0 & 25.4 & 49.6 & 61.4 & 106.1 & 3.5 & 6.1 & 11.7 & 15.2 & 53.9 & 7.4 & \textbf{15.1} & 31.7 & 42.2 & 152.4 & 16.9 & 34.4 & 76.3 & 96.8 & 164.6 \cr
LDRGCN & 13.1 & 22.0 & 37.2 & 55.8 & 97.7 & 3.4 & 6.2 & \textbf{11.2} & \textbf{13.8} & 47.3 & \textbf{6.8} & 16.3 & \textbf{27.9} & \textbf{38.9} & 131.8 & 13.2 & 32.7 & 65.1 & 91.3 & 153.5\cr
Ours & \textbf{9.1} & \textbf{16.6} & \textbf{34.7} & \textbf{44.5} & 108.4 & \textbf{3.3} & \textbf{5.9} & \textbf{11.5} & \textbf{14.7} & \textbf{44.7} & \textbf{19.6} & \textbf{31.3} & \textbf{54.8}& \textbf{66.1}& \textbf{155.5} & \textbf{12.5}& \textbf{22.7} & \textbf{44.4}& \textbf{55.8} & \textbf{120.4}\cr
& $\boldsymbol{\std{0.7}}$ & $\boldsymbol{\std{1.5}}$ & $\boldsymbol{\std{3.5}}$ & $\boldsymbol{\std{4.4}}$ & $\std{5.1}$ & $\boldsymbol{\std{1.1}}$ & $\boldsymbol{\std{2.0}}$ & $\boldsymbol{\std{3.7}}$ & $\boldsymbol{\std{4.7}}$ &$\boldsymbol{\std{15.0}}$& $\boldsymbol{\std{16.7}}$ & $\boldsymbol{\std{23.2}}$ & $\boldsymbol{\std{34.4}}$ & $\boldsymbol{\std{37.4}}$ & $\boldsymbol{\std{52.1}}$& $\boldsymbol{\std{2.0}}$ & $\boldsymbol{\std{3.8}}$ & $\boldsymbol{\std{7.4}}$ & $\boldsymbol{\std{9.6}}$ & $\boldsymbol{\std{21.0}}$    \cr \hline \hline

& \multicolumn{5}{c|}{Running} & \multicolumn{5}{c|}{Soccer} & \multicolumn{5}{c|}{Walking} & \multicolumn{5}{c}{Wash window} \cr
millisecond (ms) & 80 & 160 & 320 & 400 & 1000 & 80 & 160 & 320 & 400 & 1000 & 80 & 160 & 320 & 400 & 1000 & 80 & 160 & 320 & 400 & 1000 \cr \hline
Zero velocity& 30.6& 52.8 & 94.1 & 112.2 &242.6 & 10.3 & 17.5 & 31.8 & 39.0 & 79.4 & 18.3 & 31.2 & 55.1 &66.2 &137.7 &12.3 & 21.1&37.8 &45.7 &90.9\cr
Residual sup. & 15.6 & 19.4 & 31.2 & 36.2 & 43.3 & 20.3 & 39.5 & 71.3 & 84.0 & 129.6 & 8.2 & 13.7 & 21.9 & 24.5 & \textbf{32.2} & 8.4 & 15.8 & 29.3 & 35.4 & 61.1 \cr
convSeq2Seq & 14.3 & \textbf{16.3} & \textbf{18.0} & \textbf{20.2} & \textbf{27.5} & 12.1 & 21.8 & 41.9 & 52.9 & 94.6 & 7.6 & 12.5 & 23.0 & 27.5 & 49.8 & 8.2 & 15.9 & 32.1 & 39.9 & 58.9 \cr
FC-GCN & 25.5 & 36.7 & 39.3 & 39.9 & 58.2 & 11.3 & 21.5 & 44.2 & 55.8 & 117.5 & 7.7 & 11.8 & 19.4 & 23.1 & 40.2 & 5.9 & 11.9 & 30.3 & 40.0 & 79.3 \cr
LDRGCN & 15.2 & 19.7 & 23.3 & 35.8 & 47.4 & 10.3 & 21.1 & 42.7 & 50.9 & 91.4 & \textbf{7.1} & \textbf{10.4} & \textbf{17.8} & \textbf{20.7} & 37.5 & 5.8 & 12.3 & 27.8 & 38.2 & \textbf{56.6} \cr
Ours  & \textbf{12.4}& 19.7 & 32.3 & 39.0 & 68.9 & \textbf{4.9} & \textbf{7.9} & \textbf{14.2} & \textbf{18.0} & \textbf{53.1} & 8.1& 13.6 & 22.1 & 26.1 & \textbf{32.4} &\textbf{5.5} & \textbf{9.8} & \textbf{19.2} & \textbf{24.3} & \textbf{61.3}\cr
& $\boldsymbol{\std{0.0}}$ & $\std{0.0}$ & $\std{0.0}$ & $\std{0.0}$ & $\std{0.0}$ & $\boldsymbol{\std{0.3}}$ & $\boldsymbol{\std{0.6}}$ & $\boldsymbol{\std{1.2}}$ & $\boldsymbol{\std{1.6}}$ & $\boldsymbol{\std{4.5}}$ & $\std{0.0}$ & $\std{0.0}$ & $\std{0.0}$ & $\std{0.0}$ & $\boldsymbol{\std{1.8}}$ & $\std{0.8}$ & $\boldsymbol{\std{1.4}}$ & $\boldsymbol{\std{2.5}}$ & $\boldsymbol{\std{3.1}}$ & $\boldsymbol{\std{8.4}}$ \cr \hline

\end{tabular}
\caption{\label{tab:CMU} Motion prediction results measured with eq.\ref{eq:metric} on CMU dataset. Short-term results are reported within $80$, $160$, $320$, $400$ms, and long-term in $1$s. Best results in bold while state-of-the-art best results that fit in our confidence interval are also written bold.}
\end{table*}

\subsection{Qualitative Comparison}
\begin{figure}
    \centering
    \includegraphics[width=0.8\linewidth]{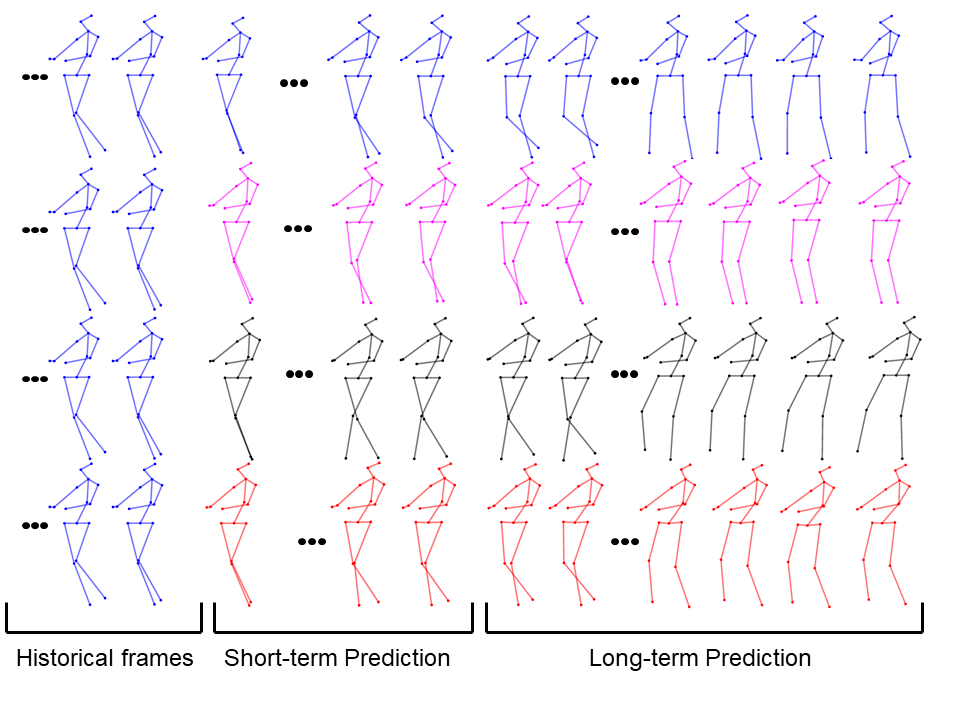}
    \caption{The left frames correspond to the sequence used as a prior. From top to bottom : ground truth, the results of ConvSeq2Seq \protect\cite{li_convolutional_2018}, FC-GCN \protect\cite{mao2020learning} and our method. The illustrated action corresponds to 'Walking Together' from Human3.6M dataset. Short-term frames shown correspond to predicted frames 1, 9 and 10 and long-term frames to frames 11, 12, 22, 23, 24 and 25.}
    \label{fig:introduction}
\end{figure}

We show in Figure~\ref{fig:introduction}, 3D pose sequences of a predicted motion using our trained model for long-term prediction. We show also the predicted 3D poses of the same sequence obtained with the baseline methods \mbox{ConvSeq2Seq}~\cite{li_convolutional_2018} and FC-GCN~\cite{mao2020learning}, based on their publicly available codes. We did not include LDRGCN \cite{Cui_2020_CVPR} in this comparison since their code is not yet available. Visually, we observe that our method produces a realistic pose sequences with a smooth motion that follows the ground truth more closely than the other methods even for long-term prediction. Our method does not show any discontinuity as a consequence of predicting the dynamic of the motion then applying it to a starting pose rather that directly predicting the pose sequence as the other methods do. 

\indent

\subsection{Smoothness of the motion}
\begin{figure}
    \centering
    \includegraphics[width=0.8\linewidth]{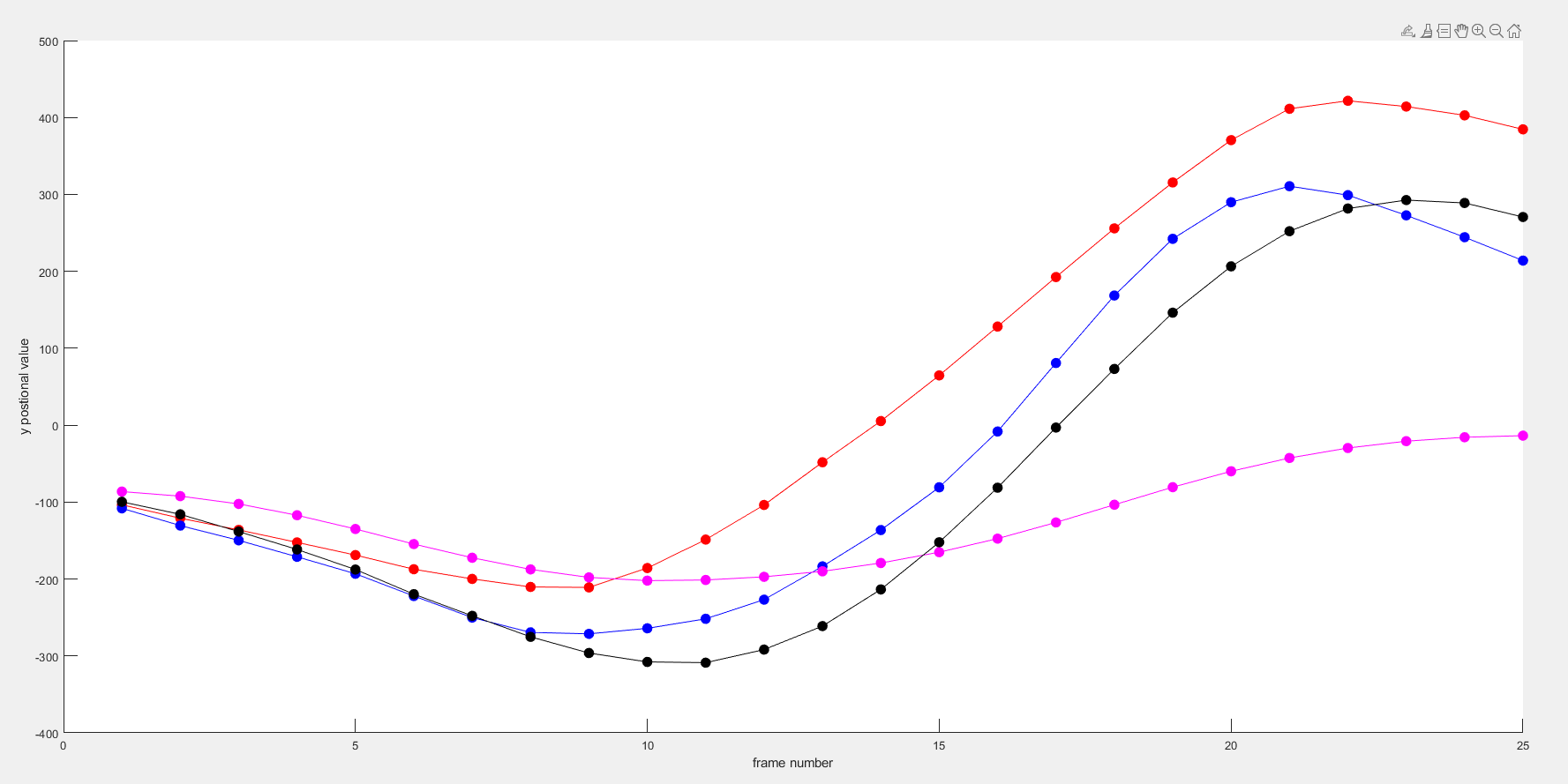}
   \caption{Walking action from Human3.6M. In blue the ground truth, in red the sequence generated by our model, in magenta ConvSeq2Seq \protect\cite{li_convolutional_2018} and in black FC-GCN \protect\cite{mao2020learning} , x-axis and y-axis corresponds respectively to frame numbers and joint position on the y axis.}
   
    \label{fig:smooth}
\end{figure}

 In order to quantitatively assess the smoothness of our predicted motions, we report in table~\ref{tab:speed}, the average euclidean distance between consecutive frames for our method against the ground truth data for some actions of the CMU MoCap dataset over all frames (25), all joints (17) and all samples from the given action (variable).  The results demonstrate that the generated movements are characterized by changes in time that are close to those shown in real videos. The Fig~\ref{fig:smooth} shows the evolution over time of the y coordinate from the skeleton's left foot on a random sample of 25 frames from the walking action from the Human3.6M dataset. 
 We represent the ground truth in blue, the sample generated by our model in red and the sample generated by ConvSeq2Seq \protect\cite{li_convolutional_2018} and FC-GCN \protect\cite{mao2020learning} in magenta and red respectively. We can see that our method produces a smooth motion that follows the motion of the ground truth. 

\begin{table}[!ht]
\centering
\small
\begin{tabular}{@{}c| c c@{} } 
 & Ground truth & Generated samples\\
 \hline
 Basketball signal & $2.33$ & $1.94$\\
 Running & $12.71$ & $10.96$\\
 Walking & $6.86$ & $6.62$\\
 Wash Window & $5.08$ & $4.24$\\
\hline
\hline
\end{tabular}
\caption{\label{tab:speed} Averaged Euclidean distance between consecutive frames for all joints and time-steps.}
\end{table}

\subsection{Computation Time}
In Table~\ref{tab:time}, we compare the computation time required by our method for long-term prediction with that of \mbox{ConvSeq2Seq} and FC-GCN. The time was obtained by predicting the long-term motion (\emph{i.e}, $25$ frames) of $8$ sequences for each of the 15 actions from Human3.6M dataset. It is worthy to note that the codes used for \mbox{ConvSeq2Seq} and FC-GCN are provided by their authors. The results of the table show that regardless of the additional computations required to map the motion back and forth to the tangent space w.r.t standard GAN models, our prediction time is similar to those of the two other methods and even faster than \mbox{ConvSeq2Seq}.

\begin{table}[!ht]
\centering
\small
\begin{tabular}{@{}c| c c@{} } 

 & total time & time per sample (25 frames)\\
 \hline
 ConvSeq2Seq & $3.04s$ & $\approx25ms$\\
 FC-GCN & $1.67s$ & $\approx14ms$\\
 Ours & $2.42s$ & $\approx20ms$\\
\hline

\hline
\end{tabular}
\caption{\label{tab:time}Prediction time comparison for $8$ predicted samples per action on Human3.6M. }
\end{table}

\subsection{Visualization}
\begin{figure*}[!ht]
\centering
\subfloat[Predicted motions]{
\label{subfig:SNEGroundTruth}
\includegraphics[width=0.4\textwidth]{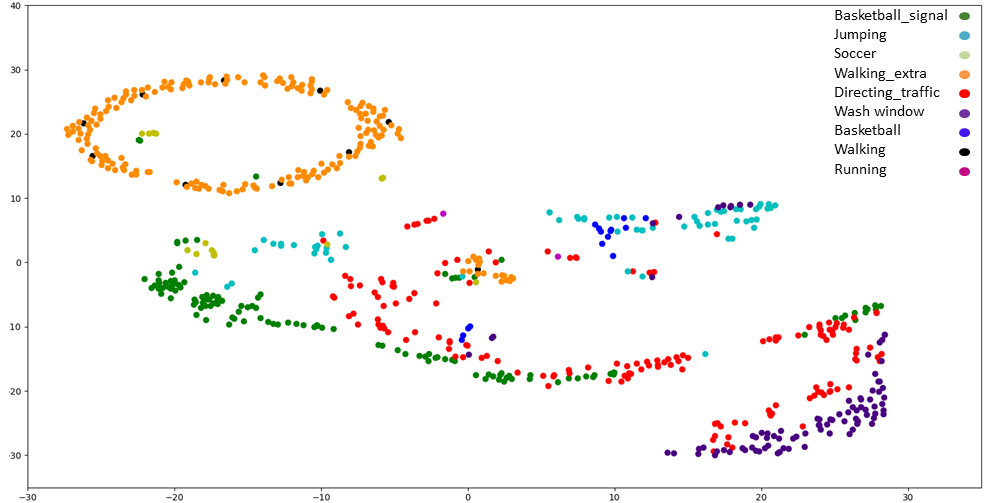}}
\subfloat[Ground truth motions]{
\label{subfig:SNEPredictiveMAWGAN}
\includegraphics[width=0.4\textwidth]{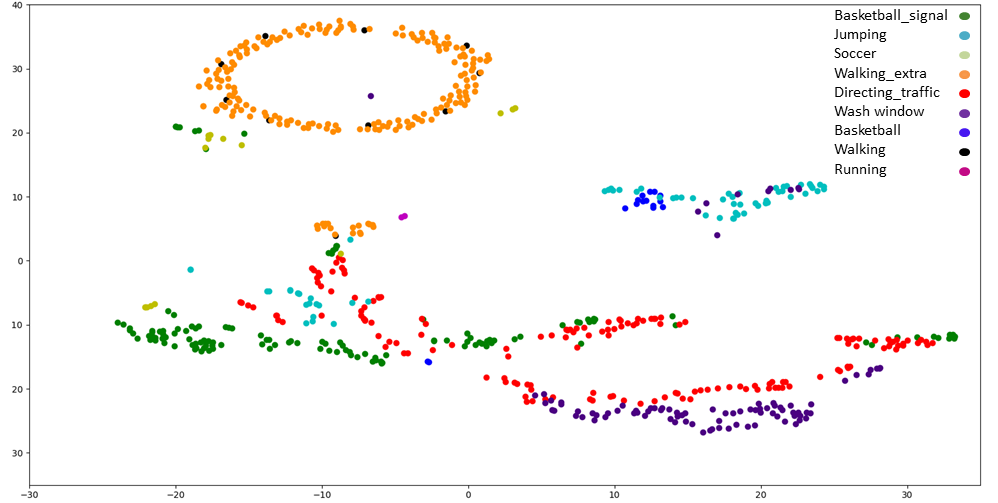} }
\caption{2D visualization of the predicted motions by our method and their associated ground truth using t-SNE algorithm based on Gram distance eq.\ref{eq:distW}. Each color represents an action.}
\label{fig:t-SNE}
\end{figure*}

To further assess the quality of the predicted samples, we present,  in Figure~\ref{fig:t-SNE} a 2D visualization of 677 long-term prediction samples from the CMU MoCap dataset predicted with our model using the t-Distributed Stochastic Neighbor Embedding (t-SNE) algorithm~\cite{vandermaaten08a}. 
This figure  clearly evidences that the predicted motions and their ground truth belong to very close distributions.  
Furthermore, the predicted 3D sequences of the same action are relatively distant from each other, what shows that our model can predict different motions for the same action while respecting the action of the given prior poses.

\subsection{Ablation Study}
In order to show the efficiency of the skeleton integrity loss $\mathcal{L}_s$ and the bone length loss $\mathcal{L}_b$ on the prediction results, we perform an ablation analysis on different models trained with different losses. We chose Human 3.6M to conduct this study motivated by the huge data it provides. In Table~\ref{tab:ablation}, we report our results for short-term and long-term  using the average error over all actions at different time steps. These results show a clear improvement when adding one of the losses, either $\mathcal{L}_s$ or $\mathcal{L}_b$, to the model that use only $\mathcal{L}_a$ and $\mathcal{L}_r$. Furthermore, while we obtain similar results for short-term prediction when using both losses (\emph{i.e.}, $\mathcal{L}_s$ and $\mathcal{L}_b$) or only $\mathcal{L}_b$, we notice a remarkable enhancement for long-term prediction  when adding both $\mathcal{L}_b$ and $\mathcal{L}_s$ to the objective function over the model that add only one of them. This evidences the importance of integrating both losses $\mathcal{L}_s$ and $\mathcal{L}_b$ to capture the spatial correlations between joints and keep predicting plausible poses in the long-term horizons.

\begin{table}[!ht]
\centering
\small
\begin{tabular}{@{}c| c c c c c@{} } 
\hline
loss functions & 80 & 160 & 320 & 400 & 1000 \\
\hline
$\mathcal{L}_a + \mathcal{L}_r$ &	20.2 &	34.9 & 	62.4 &	74.9 & 133.3\\
$\mathcal{L}_a + \mathcal{L}_r + \mathcal{L}_s$ & 13.6 &23.4& 42.6 & 51.6 & 103.8 \\
$\mathcal{L}_a + \mathcal{L}_r + \mathcal{L}_b$ & 12.6 &22.4 & 	\textbf{41.3} & 	\textbf{49.9} & 105.6\\
$\mathcal{L}_a + \mathcal{L}_r + \mathcal{L}_s + \mathcal{L}_b$ &\textbf{12.3} &	\textbf{22.2} &	\textbf{41.3} &	50.1 & \textbf{96.2} \\
 \hline
 \end{tabular}
\caption{\label{tab:ablation} Impact of the bone length loss and the skeleton integrity loss on the prediction performance for short-term and long-term. }

\end{table}
\section{Conclusions}\label{sec:conclusion}
In this paper, we have introduced a novel and robust method to deal with human motion prediction. We have represented the temporal evolution of 3D human poses as trajectories that can be mapped to points on a hypersphere. To learn this manifold-valued representation, a manifold-aware Wasserstein GAN that captures both the temporal and the spatial dependencies involved in human motion, has been proposed. We have demonstrated through extensive experiments the robustness of our method for long-term prediction compared to recent literature. This has been confirmed also by our qualitative results that show the ability of the method to produce smooth motions and plausible poses in long-term horizons.

\section{Acknowledgments}
This project has received financial support from the CNRS through the 80|Prime  program and from the French State, managed by the National Agency for Research (ANR) under the Investments for the future program with reference ANR-16-IDEX-0004 ULNE.

\section{Appendix}\label{sec:appendix}

In  this  appendix,  we  report  further experimental details. Then we  present  more visualizations and animated videos that evidence the effectiveness of the proposed approach. We also provide the code and the pre-trained models at the following \href{https://drive.google.com/drive/folders/1pQkwtVDBeubW1oPwuXWFOOtftHabKaph}{\textcolor{magenta}{link}}.

 \subsection{Experimental Details }
 In this section, we report further details about the data pre-processing for Human3.6M \cite{ionescu_human36m_2014} and CMU MoCap \footnote{\href{http://mocap.cs.cmu.edu}{http://mocap.cs.cmu.edu}} and the experiments conducted. \\
\textbf{Human3.6M. }For Human3.6M we use the database processed by \cite{jain_structural-rnn_2016} in exponential map format and use their code to convert it into 3D coordinates. We then preprocess it with a down sampling by two, from 50 fps to 25 fps, and a normalization, by subtracting the mean, dividing by the norm and subtracting the coordinates of the hips joint. Originally the dataset contains 2 long sequences for each action class and each subject. We cut those long sequences into sequences of 60 or 75 frames for short-term and long-term-prediction respectively, following \cite{li_convolutional_2018}. When creating the smaller sequences we avoid overlap, \emph{e.g.}, for sequences of 60 frames, the first sequence contains the frames 1 to 60, the second the frames 61 to 120 and so on. This leads to 3480 training samples and 812 testing samples for short-term prediction and 2769 training samples and 644 testing samples for long-term prediction.\\
 \textbf{CMU-MoCap.} We use the database proposed by \cite{li_convolutional_2018} which only contains samples from the actions basketball', 'basketball signal', 'directing traffic','jumping', 'running', 'soccer', 'walking' and 'washing window' in exponential map format. We perform the same transformation to 3D coordinate, downsampling, and normalization as for Human3.6M. We obtain our joint representation by removing the joints corresponding to duplicates joints, hands, feet and the top of the head, leading to a configuration similar to Human3.6M with 17 joints. Like Human3.6M the dataset contains several files of long motion sequences that we cut into 60 or 75 frames ones. Here, due to the small size of this dataset we have a 45 frames overlap: the first sequence contains frames 1 to 60 and the second frames 16 to 75. This leads to 2871 training samples and 704 test samples for short-term prediction and 2825 training samples and 677 test samples for long-term prediction.

\subsection{Supplementary Qualitative Results}
 
We present in figure~\ref{fig:qualitative HUMAN} two more visualizations for our generated sequences on Human3.6M: action "Walking Together", using the same sample as figure 2 from our paper and action "Phoning". We show also the predicted 3D poses of the same sequences obtained with \mbox{ConvSeq2Seq}~\cite{li_convolutional_2018} and FC-GCN~\cite{mao2020learning} as well as the ground truth. We can observe that our method produce realistic motions that follow the ground truth better than the other methods.

In figure~\ref{fig:qualitative CMU} we show a sample from CMU-MoCap for the "basketball signal" action with the corresponding ground truth in blue and our prediction in red. We see that PredictiveMA-WGAN is able to predict accurately the motion of both arms even for long-term prediction.

\subsection{Video Supplementary Materials}
\begin{figure*}[!ht]
    \centering
    \includegraphics[width=0.9\linewidth]{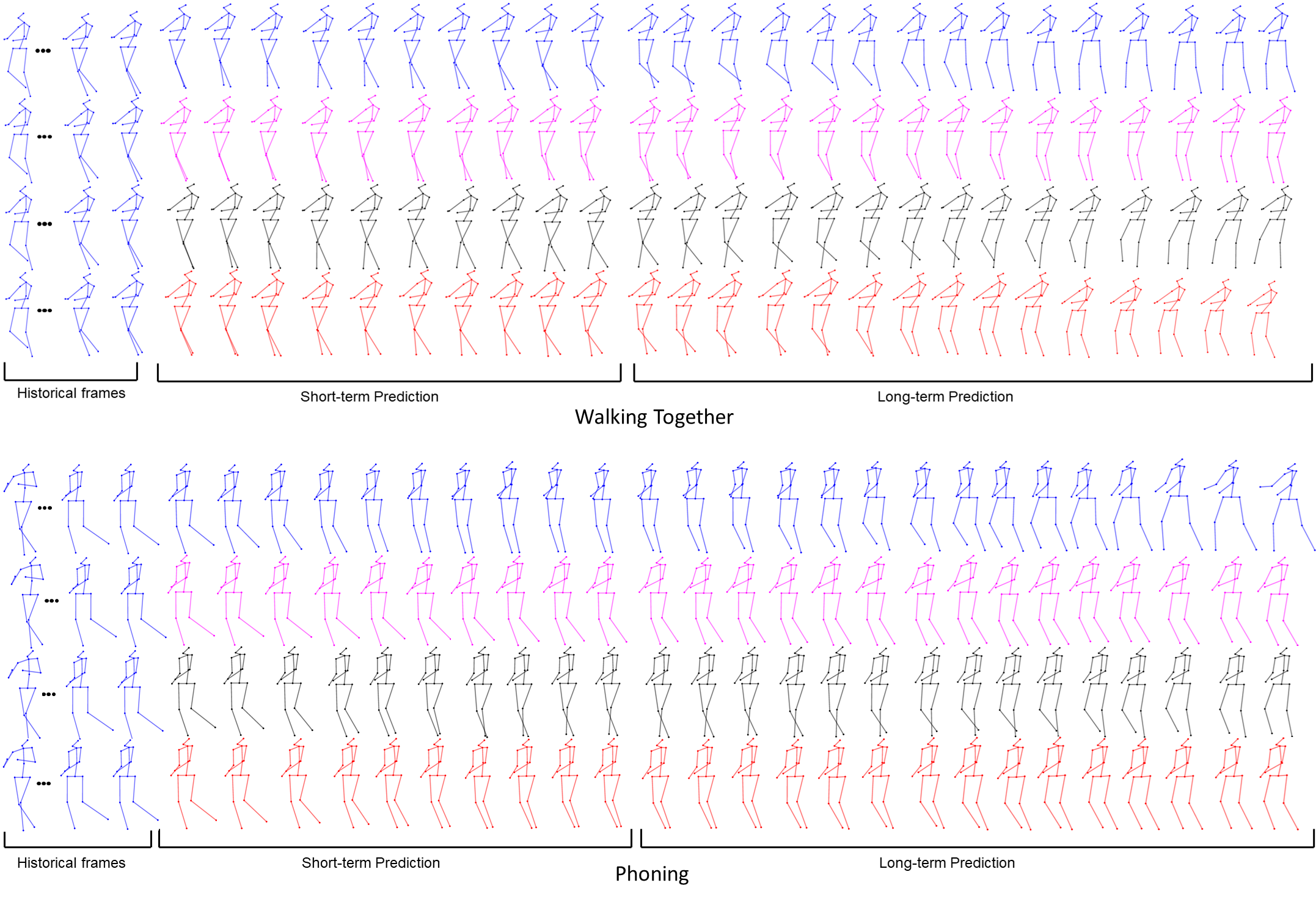}
    \caption{Qualitative comparison on the 'Walking Together' and 'Phoning' actions of Human 3.6M for long-term prediction. From top to bottom, the ground truth, the results of ConvSeq2Seq \protect\cite{li_convolutional_2018}, FC-GCN \protect\cite{mao2020learning} and our model. We can see that our method produces a smooth motion and plausible poses that better follow the ground truth.}
    \label{fig:qualitative HUMAN}
\end{figure*}

We present several videos to show the performance of our method for long-term prediction on both Human3.6M  and CMU MoCap.\\
\textbf{Human 3.6M}. On Human3.6M we show results for actions "Walking", "Walking Together", "Phoning" and "Directions" in the videos named "Human\textunderscore action-name.avi". The sequences for "Walking Together" and "Phoning" correspond to the ones shown in figure~\ref{fig:qualitative HUMAN}. In all videos we show, from left to right and top to bottom, the ground truth, our method, ConvSeq2Seq \cite{li_convolutional_2018} and FC-GCN \cite{mao2020learning}. We first show 15 historical frames for the four sequences in blue and then 25 predicted frames in blue, red, cyan and black for the ground truth, our method, \mbox{ConvSeq2Seq} and FC-GCN respectively. The videos have a frame rate of 15 fps.\\
For the "Walking" action, we can see that we follow the motion of the ground truth closely in a smooth manner for both the legs and the arms. The same observations can be made for "Walking Together". In the "Phoning" video we bring the reader attention on the motion of the legs which our model is able to accurately predict despite the irregular speed of the movement. For "Directions", we observe that our method produce the correct arm motion when the two other methods barely move.\\

\begin{figure*}[!ht]
    \centering
    \includegraphics[width=0.9\linewidth]{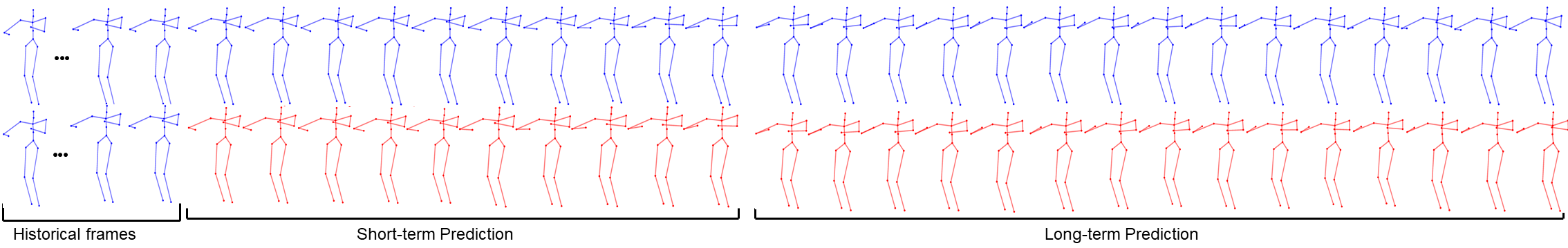}
    \caption{Qualitative comparison on the 'basketball signal' action of CMU Mocap for long-term prediction. Top: ground truth, bottom: PredictiveMA-WGAN. We predict accurately the arms motion for the entire sequence length.}
    \label{fig:qualitative CMU}
\end{figure*}

\textbf{CMU-MoCap}. We also present four videos of long-term motion prediction on the CMU-MoCap dataset for actions "Basketball", "Basketball signal", "Running" and "Wash window" in video files named "CMU\textunderscore action-name.avi". The "Basketball signal" sequence correspond to the one used in figure~\ref{fig:qualitative CMU}. We only show the ground truth and our method and follow the same representation as for Human3.6M: 15 historical frames in blue, 25 frames of prediction in blue for the ground truth and in red for ours and a frame rate of 15 fps.\\
The action "Basketball" shows that our method is able to model accurately the motion for several part of the body at the same time, here the leg and the arm follow two very different movements but the prediction stay accurate. In "Basketball signal" our model is able to predict the motion of both arms with the correct timing. For "Running" and "Wash window" we make predictions close to the ground truth for complex motions with high amplitude, especially for "Wash window" where the upper body moves in an unusual manner compared to the rest of the dataset.

\begin{thebibliography}{10}\itemsep=-1pt

\bibitem{barsoum2018hp}
E.~Barsoum, J.~Kender, and Z.~Liu.
\newblock Hp-gan: Probabilistic 3{D} human motion prediction via gan.
\newblock In {\em CVPR Workshops}, pages 1418--1427, 2018.

\bibitem{Boulbaba2016PAMI}
B.~{Ben Amor}, J.~Su, and A.~Srivastava.
\newblock Action recognition using rate-invariant analysis of skeletal shape
  trajectories.
\newblock {\em PAMI}, 38(1):1--13, 2016.

\bibitem{butepage2017deep}
J.~Butepage, M.~J. Black, D.~Kragic, and H.~Kjellstrom.
\newblock Deep representation learning for human motion prediction and
  classification.
\newblock In {\em CVPR}, pages 6158--6166, 2017.

\bibitem{Cui_2020_CVPR}
Q.~Cui, H.~Sun, and F.~Yang.
\newblock Learning dynamic relationships for 3{D} human motion prediction.
\newblock In {\em CVPR}, 2020.

\bibitem{devanne20143}
M.~Devanne, H.~Wannous, S.~Berretti, P.~Pala, M.~Daoudi, and A.~Del~Bimbo.
\newblock 3-{D} human action recognition by shape analysis of motion
  trajectories on {R}iemannian manifold.
\newblock {\em IEEE TC}, 45(7):1340--1352, 2014.

\bibitem{drira20133d}
H.~Drira, B.~{Ben Amor}, A.~Srivastava, M.~Daoudi, and R.~Slama.
\newblock 3{D} face recognition under expressions, occlusions, and pose
  variations.
\newblock {\em PAMI}, 35(9):2270--2283, 2013.

\bibitem{DBLP:conf/iccv/FragkiadakiLFM15}
K.~Fragkiadaki, S.~Levine, P.~Felsen, and J.~Malik.
\newblock Recurrent network models for human dynamics.
\newblock In {\em ICCV}, pages 4346--4354, 2015.

\bibitem{ghosh2017learning}
P.~Ghosh, J.~Song, E.~Aksan, and O.~Hilliges.
\newblock Learning human motion models for long-term predictions.
\newblock In {\em 3DV}, pages 458--466, 2017.

\bibitem{GoluVanl96}
G.~H. Golub and C.~F. Van~Loan.
\newblock {\em Matrix Computations}.
\newblock The Johns Hopkins University Press, fourth edition, 1996.

\bibitem{ferrari_adversarial_2018}
L.-Y. Gui, Y.-X. Wang, X.~Liang, and J.~M.~F. Moura.
\newblock Adversarial {Geometry}-{Aware} {Human} {Motion} {Prediction}.
\newblock In {\em ECCV}, pages 823--842. 2018.

\bibitem{gulrajani2017improved}
I.~Gulrajani, F.~Ahmed, M.~Arjovsky, V.~Dumoulin, and A.~C. Courville.
\newblock Improved training of {W}asserstein {GAN}s.
\newblock In {\em NIPS}, pages 5767--5777, 2017.

\bibitem{ZhiwuHuang2017}
Z.~Huang, J.~Wu, and L.~V. Gool.
\newblock Manifold-valued image generation with {W}asserstein generative
  adversarial nets.
\newblock In {\em {AAAI}}, pages 3886--3893, 2019.

\bibitem{ionescu_human36m_2014}
C.~Ionescu, D.~Papava, V.~Olaru, and C.~Sminchisescu.
\newblock Human3.{6M}: {Large} {Scale} {Datasets} and {Predictive} {Methods}
  for {3D} {Human} {Sensing} in {Natural} {Environments}.
\newblock {\em PAMI}, 36(7):1325--1339, July 2014.

\bibitem{jain_structural-rnn_2016}
A.~Jain, A.~R. Zamir, S.~Savarese, and A.~Saxena.
\newblock Structural-{RNN}: Deep learning on spatio-temporal graphs.
\newblock In {\em CVPR}, pages 5308--5317, 2016.

\bibitem{KacemPAMI2020}
A.~Kacem, M.~Daoudi, B.~{Ben Amor}, S.~Berretti, and J.~C.~{\'{A}}. Paiva.
\newblock A novel geometric framework on {Gram} matrix trajectories for human
  behavior understanding.
\newblock {\em PAMI}, 42(1):1--14, 2020.

\bibitem{karcher1977riemannian}
H.~Karcher.
\newblock Riemannian center of mass and mollifier smoothing.
\newblock {\em Communications on pure and applied mathematics}, 30(5):509--541,
  1977.

\bibitem{KingmaICLR14}
D.~P. Kingma and J.~Ba.
\newblock Adam: {A} method for stochastic optimization.
\newblock In {\em ICLR}, 2015.

\bibitem{KoppulaIEEEROS2013}
H.~S. {Koppula} and A.~{Saxena}.
\newblock Anticipating human activities for reactive robotic response.
\newblock In {\em IROS}, pages 2071--2071, 2013.

\bibitem{kovar2008motion}
L.~Kovar, M.~Gleicher, and F.~H. Pighin.
\newblock Motion graphs.
\newblock In {\em {ACM} {SIGGRAPH} Classes}, pages 51:1--51:10, 2008.

\bibitem{li_convolutional_2018}
C.~Li, Z.~Zhang, W.~S. Lee, and G.~H. Lee.
\newblock Convolutional {Sequence} to {Sequence} {Model} for {Human}
  {Dynamics}.
\newblock In {\em CVPR}, pages 5226--5234, 2018.

\bibitem{MaoICCV19}
W.~Mao, M.~Liu, M.~Salzmann, and H.~Li.
\newblock Learning trajectory dependencies for human motion prediction.
\newblock In {\em ICCV}, pages 9488--9496, 2019.

\bibitem{martinez_human_2017}
J.~Martinez, M.~J. Black, and J.~Romero.
\newblock On human motion prediction using recurrent neural networks.
\newblock In {\em CVPR}, pages 4674--4683, 2017.

\bibitem{OtberdoutPAMI2020}
N.~Otberdout, M.~Daoudi, A.~Kacem, L.~Ballihi, and S.~Berretti.
\newblock Dynamic facial expression generation on hilbert hypersphere with
  conditional {W}asserstein generative adversarial nets.
\newblock {\em PAMI}, pages 1--1, 2020.

\bibitem{DBLP:journals/tiv/PadenCYYF16}
B.~Paden, M.~C{\'{a}}p, S.~Z. Yong, D.~S. Yershov, and E.~Frazzoli.
\newblock A survey of motion planning and control techniques for self-driving
  urban vehicles.
\newblock {\em T-IV}, 1(1):33--55, 2016.

\bibitem{SrivastavaKJJ11}
A.~Srivastava, E.~Klassen, S.~H. Joshi, and I.~H. Jermyn.
\newblock Shape analysis of elastic curves in euclidean spaces.
\newblock {\em PAMI}, 33(7):1415--1428, 2011.

\bibitem{Turagacvpr2009}
P.~K. Turaga and R.~Chellappa.
\newblock Locally time-invariant models of human activities using trajectories
  on the {G}rassmannian.
\newblock In {\em CVPR}, pages 2435--2441, 2009.

\bibitem{vandermaaten08a}
L.~van~der Maaten and G.~Hinton.
\newblock Visualizing data using t-{SNE}.
\newblock {\em JMLR}, 9(86):2579--2605, 2008.

\bibitem{mao2020learning}
M.~Wei, L.~Miaomiao, S.~Mathieu, and L.~Hongdong.
\newblock Learning trajectory dependencies for human motion prediction.
\newblock In {\em ICCV}, pages 9488--9496, 2019.

\end{thebibliography}

\end{document}